\documentclass{article} % For LaTeX2e
\usepackage{iclr2026_conference,times}

\usepackage{hyperref}
\usepackage{url}
\usepackage{textcomp}
\usepackage[table, dvipsnames]{xcolor}
\usepackage{booktabs,multirow,array}
\usepackage[T1]{fontenc}
\usepackage{amsmath}
\usepackage{amsfonts}
\usepackage{bm}
\usepackage{xspace}
\usepackage{physics}
\usepackage{subcaption}
\usepackage{mathtools}
\usepackage{pifont}
\newcommand{\cmark}{\ding{51}}%
\newcommand{\xmark}{\ding{55}}%
\usepackage{gensymb}
\usepackage{silence}
\newcommand{\normaldist}[2]{\mathcal{N}\left( #1, #2 \right)}
\newcommand{\standardnormaldist}{\normaldist{\rv{0}}{I}}
\newcommand{\normalpdf}[2]{\varphi_{#1,#2}}
\newcommand{\pc}{\rv{x}}
\newcommand{\pcdataset}{\mathcal{X}}
\newcommand{\transspace}{\mathcal{M}}
\newcommand{\prior}{\rv{z}}
\newcommand{\partj}{\rv{p}_j}
\newcommand{\fdj}{\rv{d}_j}
\newcommand{\transj}{T_j}
\newcommand{\partembj}{\rv{w}_j}
\newcommand{\symmetryj}{\mathcal{S}_j}
\newcommand{\rv}[1]{{\boldsymbol{\mathbf{#1}}}}
\newcommand{\symmetrydist}{p_\rv{\zeta}\left( \mathcal{S}_j \mid \prior \right)}
\newcommand{\symmetrydiffusion}{s_\rv{\zeta}\left( \mathcal{S}, t \right)}
\newcommand{\pcenc}{q_{\rv{\eta}}\left(\prior \mid \pc \right)}
\newcommand{\pcdec}{p_{\rv{\eta}}\left(\pc \mid \prior \right)}
\newcommand{\vaeprior}{p(\prior)}
\newcommand{\priordiffusion}{p_{\rv{\theta}}\left(\prior\right)}
\newcommand{\fddiffusion}{p_{\rv{\xi}}\left(\fdj \mid \symmetryj, \prior\right)}
\newcommand{\partdist}{p_{\rv{\xi}}\left(\partj \mid \symmetryj, \prior\right)}
\newcommand{\partenc}{q_{\rv{\phi}}\left(\partembj \mid \partj, \prior\right)}

\newcommand{\assemblerdist}{p_{\rv{\psi}}\left(\transj \mid \partembj, \partj, \symmetryj, \prior\right)}
\newcommand{\assemblerdiffusion}{p_{\rv{\psi}}\left(\transj \mid \partembj, \prior\right)}
\newcommand{\pcdist}{p\left(\pc \right)}
\newcommand{\oursfull}{Quartet of Diffusions\xspace}
\newcommand{\ours}{Quartet\xspace}
\AtEndPreamble{
    \usepackage[capitalize]{cleveref}
    \crefname{section}{Sec.}{Secs.}
    \Crefname{section}{Section}{Sections}
    \Crefname{table}{Table}{Tables}
    \crefname{table}{Tab.}{Tabs.}
    \Crefname{algorithm}{Algorithm}{Algorithms}
    \crefname{algorithm}{Alg.}{Algs.}
    \Crefname{figure}{Figure}{Figures}
    \crefname{figure}{Fig.}{Figs.}
}

\RequirePackage{enumitem}
\setlist[itemize]{noitemsep,leftmargin=*,topsep=0em}
\setlist[enumerate]{noitemsep,leftmargin=*,topsep=0em}

\makeatletter
\DeclareRobustCommand\onedot{\futurelet\@let@token\@onedot}
\def\@onedot{\ifx\@let@token.\else.\null\fi\xspace}

\def\eg{\emph{e.g}\onedot} 
\def\ie{\emph{i.e}\onedot} 
 
 \def\vs{\emph{vs}\onedot}

\def\etal{\emph{et al}\onedot}
\makeatother

\let\titleold\title
\renewcommand{\title}[1]{\titleold{#1}\newcommand{\thetitle}{#1}}
\def\maketitlesupplementary
   {{
        \centering
        \Large
        \textbf{\thetitle}\\
        \vspace{0.5em}Appendix \\
        \vspace{1.0em}
   }}

\renewcommand{\cite}{\citep}

\title{\oursfull: Structure-Aware Point Cloud Generation through Part and Symmetry Guidance}

\author{Chenliang Zhou, Fangcheng Zhong, Weihao Xia, Albert Miao, Canberk Baykal, Cengiz Oztireli
\\
Department of Computer Science and Technology, University of Cambridge\\
\texttt{chenliang.zhou@cst.cam.ac.uk} \\
}

\iclrfinalcopy % Uncomment for camera-ready version, but NOT for submission.

\usepackage{graphicx}
\usepackage{etoolbox}
\newcommand{\insertfig}{\includegraphics[width=\linewidth]{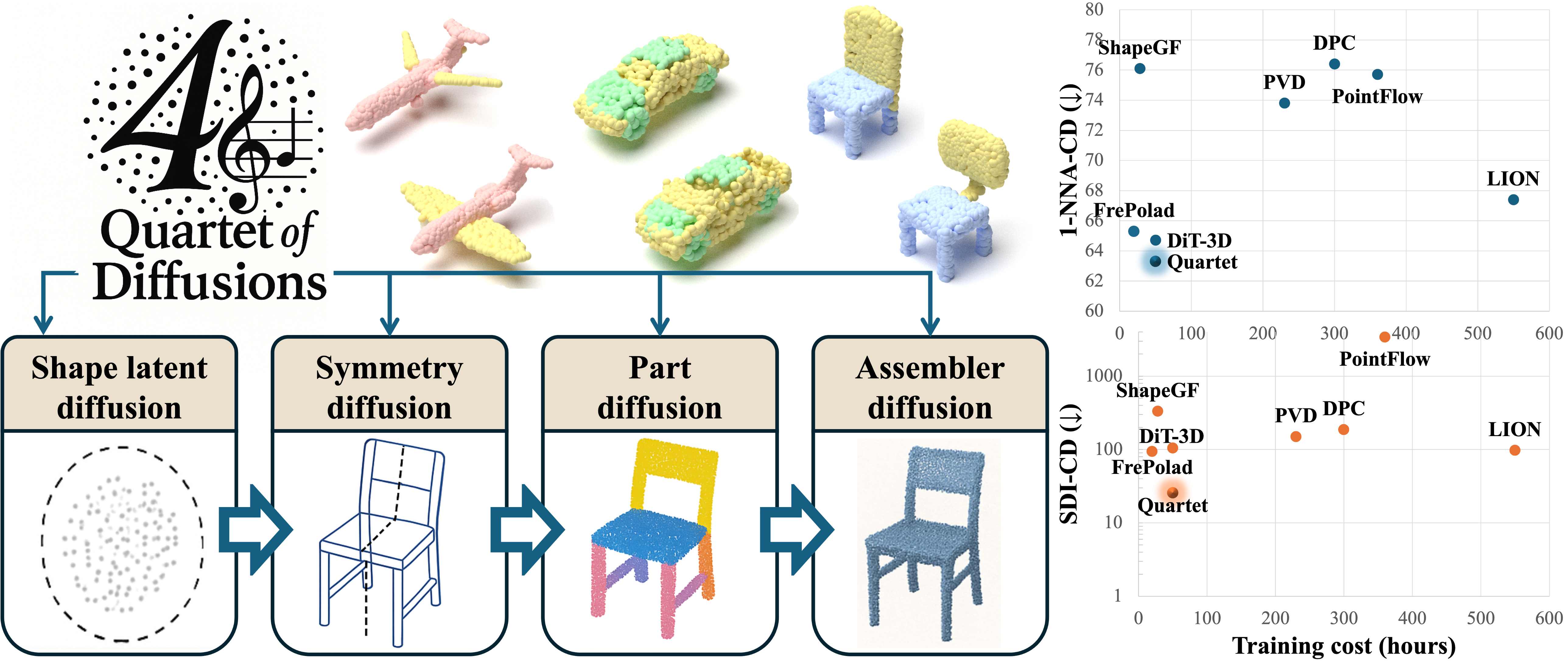}\captionof{figure}{Our method, the \oursfull, harmoniously orchestrates an interpretable and controllable pipeline for structure-aware, high-quality, diverse, and efficient point cloud generation guided by part and symmetry clues. 
}
\label{fig:teaser}}% define the image

\makeatletter
\apptocmd{\@maketitle}{\centering\insertfig\medskip}{}{}% insert the figure after authors
\makeatother
\begin{document}
\maketitle
\begin{abstract}
We introduce the \emph{\oursfull}, a structure-aware point cloud generation framework that explicitly models part composition and symmetry. Unlike prior methods that treat shape generation as a holistic process or only support part composition, our approach leverages four coordinated diffusion models to learn distributions of global shape latents, symmetries, semantic parts, and their spatial assembly. This structured pipeline ensures guaranteed symmetry, coherent part placement, and diverse, high-quality outputs. By disentangling the generative process into interpretable components, our method supports fine-grained control over shape attributes, enabling targeted manipulation of individual parts while preserving global structural consistency. A central global latent further reinforces structural coherence across assembled parts. Our experiments show that the Quartet achieves state-of-the-art performance in generating high-quality and diverse point clouds while maintaining symmetry. To our best knowledge, this is the first 3D point cloud generation framework that fully integrates and enforces both symmetry and part priors throughout the generative process.
\end{abstract}

\section{Introduction}
\label{sec:introduction}
Structure-aware 3D shape generation aims to model not only the holistic surface geometry of objects but also the \textbf{underlying organizational principles} that govern their composition and formation~\cite{chaudhuri2020learning}. A truly structure-aware representation must comprise two essential components~\cite{chaudhuri2020learning}: the geometry of atomic structural elements (\eg, low-level parts) and the structural patterns that dictate how these elements are assembled into coherent shapes or scenes. These patterns may manifest as hierarchical decompositions, part-whole relationships, spatial composition graphs, or local and global symmetries.

In this work, we focus on 3D point cloud generation with \textbf{explicit awareness of two fundamental structural priors} commonly observed in natural and man-made objects: \textbf{part awareness} and \textbf{symmetry awareness}. \emph{Part awareness} captures the notion that objects are composed of semantically meaningful, interrelated components with geometrically distinct properties. \emph{Symmetry awareness}, in contrast, models the intrinsic geometric regularities—such as reflective or rotational symmetries—that often govern the spatial arrangement of parts. Importantly, part awareness serves as a \emph{prerequisite} for identifying and enforcing local symmetries, which typically operate at the level of inter-part relationships. Without symmetry, shapes risk appearing disjointed or unnatural, undermining their functional performance in practical applications
(\eg some prior work in \cref{fig:exp-gen}).
% (\eg some objects generated by prior work in \cref{fig:exp-gen}).

% \fz{Symmetry plays a pivotal role in 3D shape generation as it inherently reflects the regularities and predictable patterns that govern the structure of both natural and man-made objects.
% Examples of strict symmetry include rotational symmetry in gears, chandeliers, or wheelchairs; reflective symmetry in a butterfly’s wings, aircraft, or automobiles; and translational symmetry in tiled floor patterns or brick walls.
% These symmetries serve as key constraints ensuring that generated objects maintain spatial coherence, exhibit geometrically consistent parts, and adhere to natural or engineered laws.
% This not only improves the visual realism of the generated shapes but also enhances their structural integrity.
% Without symmetry, shapes risk appearing disjointed or unnatural, undermining functional performance in practical applications.
% }

Compared to unstructured 3D data, structured representations offer substantial advantages. They not only enhance the \emph{visual plausibility} and \emph{structural coherence} of generated shapes, but also facilitate downstream tasks such as part segmentation, alignment, structure inference, and fine-grained shape editing~\cite{schor2019componet, li2020learning, hertz2022spaghetti, koo2023salad}. However, existing generative approaches largely remain structure-oblivious -- treating point cloud synthesis as an unconstrained distribution learning problem. These methods often prioritize global visual fidelity and geometry-level details in the distribution, while neglecting the explicit modeling of inherent structure~\cite{vahdat2022lion, yang2019pointflow, mo2023dit, zhou20213d}. As a result, they tend to produce shapes that suffer from poor part organization, broken or inconsistent symmetries, and limited generalization to unseen or complex object categories~\cite{li2020learning}. While a few recent efforts support part-level generation, they rarely incorporate symmetry as a \emph{learned}, \emph{relational prior}. To date, there remains a notable gap: the absence of unified, principled frameworks that jointly encode \emph{part-level composition} and \emph{symmetry structures} to guide the generation of coherent, interpretable, and generalizable 3D shapes.

To address this, we propose the \emph{\oursfull} (the \emph{\ours}), a structure-aware point cloud generation framework that explicitly leverages part and symmetry guidance. Specifically, our pipeline employs four diffusion models~\cite{ho2020denoising} to learn the distributions of shape latents, symmetries, semantic parts, and assemblers that assemble the full point cloud from parts. By explicitly modeling these structural distributions, the ensemble of four diffusions harmoniously orchestrates an effective pipeline for structure-aware point cloud generation through part and symmetry guidance: Our method enables the generation of high-quality, diverse point clouds with guaranteed symmetry. The disentangled structural representation makes the generation process interpretable and controllable, facilitating targeted modifications to individual parts while preserving global consistency. Structural coherence is further reinforced by a global shape latent, which anchors part assembly to the overall geometry. This approach addresses key limitations in structure modeling and represents the first 3D shape generation method to guarantee symmetry in the generated shapes.

\section{Related Work}
\paragraph{Symmetry in 3D shapes} Symmetry is a fundamental geometric property observed across natural and human-made objects. Extensive research exists on symmetry detection~\cite{mitra2013symmetry, atallah1985symmetry, illingworth1988survey, mitra2006partial, je2024robust, zhou2021nerd, gao2020prs, fukunaga1975estimation, comaniciu2002mean} (for more details, see \cref{sec:preliminary-symmetry,sec:supp-symmetry-detection}). Symmetry clues are widely used in various 3D vision tasks including acquisition and representation~\cite{yang2024sym3d, buades2008nonlocal, li2010analysis, pauly2005example, thrun2005shape, xu2009partial, zheng2010non}, classification~\cite{kazhdan2004symmetry, martinet2006accurate, podolak2006planar}, perception~\cite{reisfeld1995context} , manipulation~\cite{mitra2007symmetrization, podolak2007symmetry, panozzo2012fields, gal2009iwires, mehra2009abstraction}, reconstruction~\cite{phillips2016seeing, xu2024instantmesh, tulsiani2020object}, inverse rendering~\cite{wu2020unsupervised}, and refinement~\cite{mitra2007symmetrization}. However, symmetry-aware 3D shape generation remains underexplored. Aside from our method, few approaches guarantee symmetry. PAGENet~\cite{li2020learning} promotes symmetry via an MSE loss between a shape and its reflection, but it is limited to reflectional symmetry and does not guarantee it.

\paragraph{Part-based 3D shape generation}
Part-based 3D shape generation leverages the modular structure of objects~\cite{mitra2014structure, mitra2014structure2} and emerges as a vital paradigm for modeling complex geometries~\cite{zerroug1999part, kim2013learning}, enhancing diversity~\cite{schor2019componet, chen2024partgen} and facilitating downstream tasks including recognition~\cite{hoffman1984parts}, retrieval~\cite{chang2015shapenet, mitra2014structure}, and manipulation~\cite{huang2014functional}.
% Traditional holistic shape generation techniques often suffer from limited diversity and structural plausibility, prompting researchers to explore part-based decompositions that align more naturally with human perception and design patterns.
Early methods focus on geometric template fitting or hierarchical part assembly~\cite{funkhouser2004modeling, bokeloh2010connection, kalogerakis2012probabilistic, berthelot2017began, cohen2016inspired, fish2014meta}. More recent work harnesses implicit shape representations~\cite{koo2023salad, hertz2022spaghetti, talabot2025partsdf, hui2022neural, genova2019learning, genova2020local}. 
For example, PartSDF~\cite{talabot2025partsdf} models parts using implicit neural fields, enabling both continuous interpolation and discrete composition.

Deep neural networks are widely adopted for part-based shape modeling~\cite{schor2019componet, li2020learning, dubrovina2019composite, li2024pasta, wu2020pq, huang2015analysis, li2017grass, mo2019structurenet, zou20173d, nash2017shape, wu2019sagnet, wang2018global, wang2019dynamic, gao2019sdm}. For instance, CompoNet~\cite{schor2019componet} enhances diversity by varying both parts and their compositions, while PASTA~\cite{li2024pasta} generates shapes conditioned on part arrangement for fine-grained control. Similarly, our \ours performs structure-aware 3D generation by modeling both parts and their compositions, with the added benefit of symmetry enforcement.
% while maintaining global coherence.

% Another important direction is latent space manipulation. PQ-NET (Wu et al., 2020) learned compact latent codes for shape parts and employed a VAE-based structure to explore part permutations. In contrast, 3DShape2VecSet (Zhang et al., 2022) formulated shapes as unordered part sets and learned permutation-invariant embeddings, enhancing the robustness and diversity of generated outputs.

\paragraph{Diffusion models}
Diffusion models~\cite{ho2020denoising} 
% ~\cite{sohl2015deep, ho2020denoising}
are generative models based on Markovian diffusion processes (see \cref{sec:preliminary-diffusion} for more details). They have demonstrated remarkable performance in various domains~\cite{croitoru2023diffusion, cao2024survey, yang2023diffusion}, particularly in image~\cite{dhariwal2021diffusion, ho2020denoising, rombach2022high, ramesh2022hierarchical, brooks2023instructpix2pix, saharia2022image}, speech~\cite{chen2020wavegrad, jeong2021diff, liu2022diffgan}, video~\cite{ho2022video, xing2024survey, luo2023videofusion, yang2023diffusion}, 3D scene~\cite{wei2023lego, zhai2023commonscenes, tang2024diffuscene}, and 3D object generation~\cite{zhou20213d, luo2021diffusion, vahdat2022lion, nakayama2023difffacto, wu2023fast, mo2023dit, liu2019point, zhou2024frepolad, koo2023salad}.
% Other successful applications include density estimation~\cite{kingma2021variational}, image restoration~\cite{}. 
Accordingly, our \ours recruits four diffusion models to learn the distributions of shape latents, symmetries, parts, and assemblers. 

\section{Method}
\label{sec:method}
We aim to model the distribution of point clouds $\pc \in \pcdataset \subseteq \mathbb{R}^{3 \times N}$ as that over a collection of semantic parts $\{\rv{p}_j\}_{j=1}^M$ through part and symmetry guidance. Specifically, we view each point cloud $\pc = \bigcup_{j=1}^M \transj \partj$ as a composition of its $M$ semantic parts and their corresponding assemblers.
% \begin{equation}
%     \pc = \left\{ \left( \rv{p}_0, \rv{t}_0 \right), \left( \rv{p}_1, \rv{t}_1 \right), \ldots, \left( \rv{p}_M, \rv{t}_M \right) \right\},
% \end{equation}
Each assembler $\transj$ is a transformation
% \begin{equation}
% \transj = \left(m_x, m_y, m_z, \alpha_x, \alpha_y, \alpha_z, s_x, s_y, s_z \right) \in \mathbb{R}^9
% \end{equation}
comprising translation, rotation, and scaling in 3D Euclidean space, mapping the part $\partj$ to the correct position, orientation, and scale in the original point cloud.
%$\pc$.
% by
% \begin{equation}
%     \pc = \transj\left(\partj\right) \coloneqq \rv{s}_j \odot \left(R_{z,\alpha_z} R_{y,\alpha_y}  R_{x,\alpha_x} \partj \right) + \rv{m}_j,
% \end{equation}
% where $\odot$ denotes the Hadamard product, and $R_{x, \alpha}$, $R_{y, \alpha}$, and $R_{z, \alpha}$ are rotations by the angle $\alpha$ about the $x$-, $y$-, and $z$-axes, respectively.

\begin{figure}
    \centering
    \includegraphics[width=\linewidth]{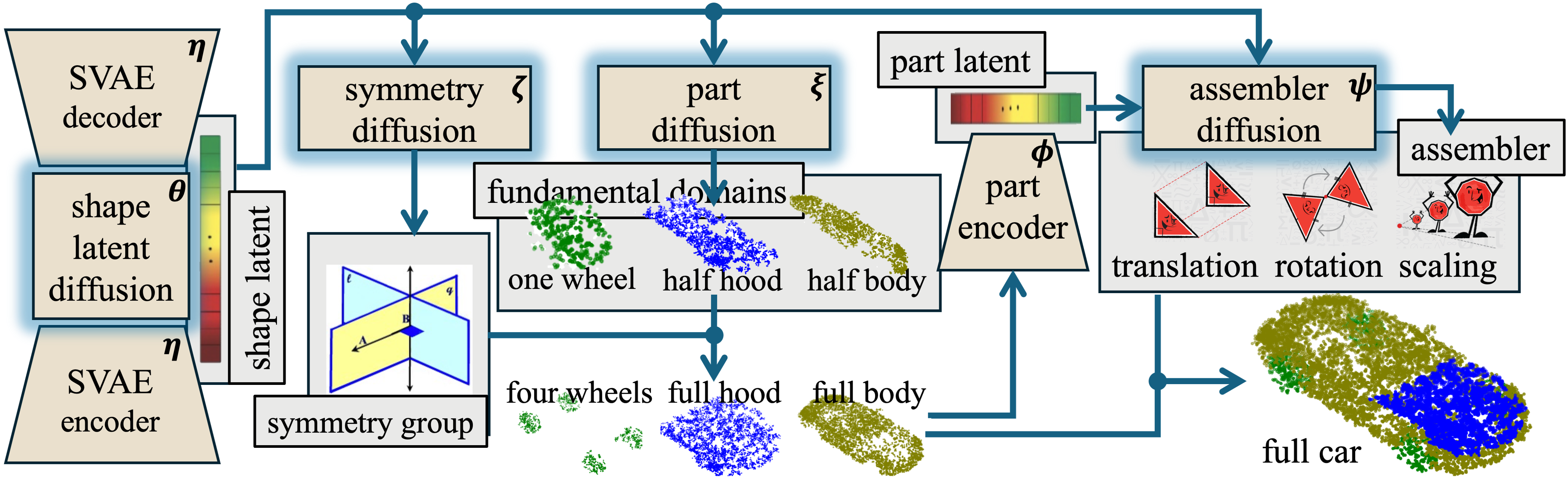}
    \caption{Overview of the \ours's architecture. Four diffusions are employed to learn the distributions of shape latents, symmetries, parts, and assemblers. By explicitly modeling different distributions, the \ours provides an interpretable and controllable framework to generate high-quality, diverse 3D shapes with guaranteed symmetry. Beige blocks denote learnable modules; gray blocks indicate outputs directly generated from the models.}
    \label{fig:overview}
\end{figure}

% \FZ{Best not introduce any variables without defining them. We may briefly explain what each module does only in words.}
The \ours models four distributions using parameterized diffusion models: 1. point cloud shape latents $\priordiffusion$ (\cref{sec:prior-distribution}), 2. part-wise symmetries $\symmetrydist$ (\cref{sec:method-symmetry-enforcement}), 3. point cloud parts $\partdist$ (\cref{sec:method-part-generation}), and 4. assemblers $\assemblerdist$, additionally conditioned on part latent $\partembj$ (\cref{sec:method-part-assembly}).
% on the corresponding transformation $\transj$ to assemble different parts while ensuring the symmetry in each part (\cref{sec:method-part-assembly}).
% Note that some distributions are also conditioned on a shape latent $\prior$ (\cref{sec:prior-distribution}) and part latents $\partembj$. 
These components collectively define the \emph{\oursfull} architecture. Pipeline overviews are shown in \cref{fig:teaser,fig:overview}.
% in modeling the point cloud shape latent $\priordiffusion$, symmetry $p_\rv{\zeta}\left( \mathcal{S}_j \mid \prior \right)$, part $\partdist$, and assembler distributions $\assemblerdist$, respectively. 
% The following four subsections will discuss these four members of the \ours in detail.

Modeling point cloud distributions in this way offers several key advantages: 1. It introduces variability at both the shape and assembly levels, enabling combinatorially greater diversity; 2. By explicitly generating and enforcing symmetry early in the pipeline, the model produces outputs with more realistic and consistent symmetric properties; 3. The separate modeling of symmetries, parts, and assembly allows for better interpretability and fine-grained control -- individual parts can be manipulated independently without compromising global structure; and 4. A central shape latent $\prior$ ensures structural coherence across all components.
These benefits are demonstrated through extensive experiments in \cref{sec:experiments}.

The \ours is trained by sequentially optimizing the parameters of four distributions to fit the given dataset $\pcdataset$.
Point cloud generation is performed in two phases: part generation and part assembly. In part generation, a shape latent $\prior$ is first sampled from the latent distribution $\priordiffusion$. Conditioned on $\prior$, symmetry groups $\symmetryj$ are sampled from $\symmetrydist$, and parts $\partj$ are drawn from $\partdist$ for $j = 1, 2, \ldots, M$; In part assembly, parts $\partj$ are encoded into latent representations $\partembj$ using an encoder $\partenc$, and the corresponding assemblers $\transj$ are sampled from $\assemblerdist$. These assemblers are then applied to the parts to reconstruct the full point cloud
\begin{equation}
    \tilde{\pc} \coloneqq \bigcup_{j=0}^M \transj \partj.
\end{equation}
Mathematically, the generation process can thus be formulated as
{\small
\begin{align}
&p\left( \left\{ \left( \partj, \transj, \symmetryj, \partembj \right)\right\}_{j=0}^M, \prior \right) 
= \priordiffusion p\left( \left\{ \left( \partj, \transj, \symmetryj, \partembj \right)\right\}_{j=0}^M \,\middle\vert\, \prior \right)
= \priordiffusion \prod_{j=0}^M p(\partj, \transj, \partembj \mid \prior) \label{eq:gen-dist-middle} \\ 
&= \priordiffusion \prod_{j=0}^M \symmetrydist \partdist \partenc \assemblerdist \label{eq:gen-dist},
\end{align}
}
\hspace{-0.5em}
where the second equality in (\ref{eq:gen-dist-middle}) follows from the conditional independence of parts given the shape latent $\prior$.
Therefore, the overall point cloud distribution can be modeled as
{
\small
\begin{align}
&\pcdist
= p\left( \left\{ \left( \partj, \transj \right) \right\}_{j=0}^M\right) = \idotsint p\left( \left\{ \left( \partj, \transj, \symmetryj, \partembj \right)\right\}_{j=0}^M, \prior \right)  \left( \prod_{j=1}^M \dd \symmetryj \right) \left( \prod_{j=1}^M \dd \partembj \right) \dd \rv{z} \\
&\overset{(\ref{eq:gen-dist})}{=} \int \priordiffusion \prod_{j=0}^M \left( \int \int \partdist \partenc \assemblerdist \dd \symmetryj \dd \partembj \right) \dd \prior. \label{eq:pc-dist}
\end{align}
}

\subsection{Shape Latent Diffusion}
\label{sec:prior-distribution}
Although we model each distribution separately, they must remain aware of the underlying point cloud to coordinate coherently. Directly conditioning on the full point cloud is computationally expensive due to its high dimensionality. Instead, we condition them on shape latents obtained from a variational autoencoder (VAE)~\cite{kingma2013auto, zhou2024frepolad, vahdat2022lion} (see \cref{sec:preliminary-vae} for background).

To effectively encode point clouds while preserving semantic information, we propose a novel \emph{sparse variational autoencoder} (\emph{SVAE}) with a latent diffusion modeling its latent distribution. SVAE builds on a VAE architecture implemented with point-voxel convolutional neural networks (PVCNNs)~\cite{liu2019point, zhou20213d}, drawing inspiration from sparse autoencoders~\cite{ng2011sparse}. 
% ~\cite{ng2011sparse, bricken2023monosemanticity}.
Prior work suggests that enforcing sparsity in activation layers improves interpretability~\cite{cunningham2023sparse, makelov2024towards, marks2024enhancing} and semantic disentanglement~\cite{bricken2023monosemanticity, o2024disentangling}, which is critical for our downstream tasks. To this end, we introduce a sparsity constraint on the final activation layer $a_\rv{\eta}(\rv{x})$ of the encoder $\pcenc$ while maximizing the VAE evidence lower bound (ELBO) $\mathcal{L}_\text{ELBO}$ (\cref{eq:vae-elbo-no-beta} in \cref{sec:preliminary-vae}):
\begin{equation}
\begin{aligned}
\label{eq:abvae-optimization-problem}
    \underset{\rv{\eta}}{\text{max}}\; \mathbb{E}_{\rv{x}} \left[ \mathcal{L}_\text{ELBO}(\rv{\eta}; \rv{x}) \right] \; \text{subject to}\; \mathbb{E}_{\rv{x}} \left[ \left\| a_\rv{\eta}(\rv{x}) \right\|_1 \right] < \delta,
\end{aligned}
\end{equation}

\hspace{-0.5em}
where $\delta > 0$ controls the sparsity strength. Leveraging Karush–Kuhn–Tucker (KKT) methods~\cite{kuhn1951nonlinear, karush1939minima}, we reformulate the constrained optimization problem into a Lagrangian $\mathcal{F}(\rv{\eta}, \lambda; \rv{x})$ whose optimal point is a global maximum over the domain of $\rv{\eta}$ and obtain its lower bound:
\begin{align}
    \mathcal{F}(\rv{\eta}, \lambda; \rv{x}) &:= \mathcal{L}_\text{ELBO}(\rv{\eta}; \rv{x}) - \lambda \left( \left\| a_\rv{\eta}(\rv{x}) \right\|_1 - \delta \right) \\
    &\geq \mathcal{L}_\text{ELBO}(\rv{\eta}; \rv{x}) - \lambda \left\| a_\rv{\eta}(\rv{x}) \right\|_1 =: \mathcal{L}_\text{SVAE}(\rv{\eta}; \rv{x}), \label{eq:svae-loss}
\end{align}
\hspace{-0.5em}
where $\lambda$ is the KKT multiplier.
% Since $\delta > 0$, we obtain a lower-bound on the ELBO:
% \begin{equation}
% \label{eq:freelbo}
% \begin{aligned}
%      \mathcal{F}(\rv{\eta}, \lambda; \rv{x})
%     % &= \mathcal{F}_\beta(\rv{\psi}, \rv{\xi}; \rv{X}) - \eta \left( \mathcal{L}_\text{Fre} \left(\vaeenc, \vaedec; \rv{X} \right) - \delta \right) \\
%     % &= \mathcal{F}_\beta(\rv{\psi}, \rv{\xi}; \rv{X}) - \eta \mathcal{L}_\text{Fre} \left(\rv{X}, \vaeenc, \vaedec \right) + \eta \delta \\
%     \geq \mathcal{L}_\text{ELBO}(\rv{\eta}; \rv{x}) - \lambda \left\| a_\rv{\eta}(\rv{x}) \right\|_1 
%     % &= \mathbb{E}_{\vaeenc} \left( \log \vaedec \right)  - \eta \mathcal{L}_\text{Fre} \left(\vaeenc, \vaedec; \rv{X} \right) - \beta \mathcal{D}_\text{KL} \left( \vaeenc, \vaeprior; \rv{X} \right) \\
%     % & =: \mathcal{L}_\text{FreELBO}\left(\rv{\psi}, \rv{\xi}, \eta; \rv{X} \right).
% \end{aligned}
% \end{equation}
The SVAE is trained by maximizing this lower bound $\mathcal{L}_\text{SVAE}$. The hyperparameter $\lambda$ controls the trade-off between reconstruction quality and sparsity, with $\lambda = 0$ recovering the standard ELBO.

While a simple Gaussian prior is commonly used for VAE latent distributions, evidence suggests that such a restricted prior cannot accurately capture complex latent distributions (\ie the \emph{prior hole problem}~\cite{vahdat2021score, zhou2023clip}) 
% ~\cite{vahdat2021score, tomczak2018vae, rosca2018distribution, zhou2023clip})
and can degrade VAE performance~\cite{chen2016variational}. To address this, we follow prior work~\cite{zhou2024frepolad} and employ a diffusion model $\priordiffusion$, implemented with a U-Net backbone,
%~\cite{ronneberger2015u},
to learn the latent distribution. Once trained, the shape latents can be directly sampled from it. This design enables effective and interpretable feature extraction via SVAE while enhancing latent expressiveness and maintaining runtime efficiency.

\subsection{Symmetry Enforcement}
\label{sec:method-symmetry-enforcement}
\begin{figure}
    \centering
    \includegraphics[width=\linewidth]{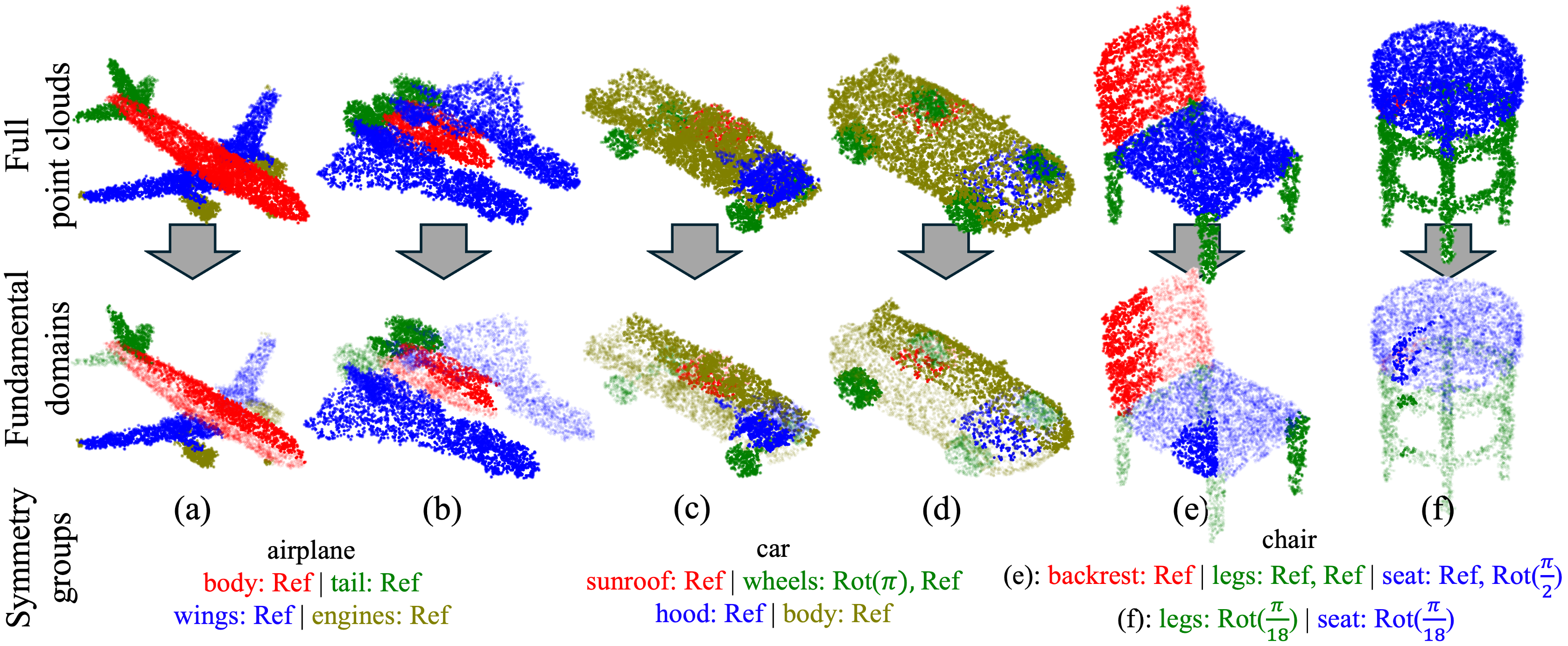}
    \caption{Point cloud airplanes, cars, and chairs with identified symmetry groups and corresponding fundamental domains for each color-coded part. Ref denotes reflection; Rot($\alpha$) denotes rotation by angle $\alpha$. Our symmetry formulation allows greater flexibility by supporting symmetries composed of multiple transformations, such as two reflections (chair legs (e)) or a rotation followed by a reflection (car wheels). Circular symmetry is approximated via small-angle rotations (chair seat (f)).}
    \label{fig:show_symmetry}
\end{figure}

The \ours guarantees symmetry by explicitly learning its distribution. Once learned, the model can generate and enforce appropriate symmetries during point cloud generation. For a 3D shape $\rv{p}$, a finite group of rigid transformations $\mathcal{S} = \langle \mathcal{T} \rangle \subseteq \mathrm{E}(3)$, generated by $\mathcal{T} = \{S_1, S_2, \ldots, S_n\}$, is said to be a \emph{symmetry group} over a subset $\rv{d} \subseteq \rv{p}$ if the full shape $\rv{p}$ can be reconstructed by $\mathcal{S} \rv{d}$, 
the application of $\mathcal{S}$ on $\rv{d}$, 
which is defined to be the union of all images under the transformations in $\mathcal{S}$:
\begin{equation}
\label{eq:symmetry-defn}
    \underbrace{\bigcup_{S \in \mathcal{S}} S \rv{d}}_{\mathcal{S} \rv{d}} = \rv{p}, 
    % \bigcup_{S \in \mathcal{S}} S \rv{d} = \rv{p}, 
\end{equation}
\hspace{-0.4em}where the group generation is defined as
\begin{equation}
    \langle \mathcal{T} \rangle \coloneqq \left\{\prod_{i=1}^{n} S_i \,\middle\vert\, S_i \in \mathcal{T}, n \in \mathbb{N} \right\}.
\end{equation}
% where $n \in \mathbb{Z}^+$ is the \emph{order of symmetry}, namely the smallest number such that $S^n=I$, the identity transformation.
The minimal such subset $\rv{d}$ is called the \emph{fundamental domain} for $\mathcal{S}$.
% , which serves as a basis for the symmetry.
\Cref{fig:show_symmetry} illustrates fundamental domains for various parts in point cloud airplanes, cars, and chairs.

As translational symmetry implies an infinite shape, for tractability we restrict symmetry groups to those generated by at most two transformations -- reflections or rotations with angles $\alpha \geq \frac{\pi}{18}$ such that $\frac{2\pi}{\alpha} \in \mathbb{Z}$. By a special case of the Cartan–Dieudonné theorem~\cite{gallier2011cartan}, any 3D rotation can be expressed as a composition of two reflections across planes intersecting along the axis of rotation, where the rotation angle is twice the angle between the two planes. Consequently, we just need to search for the symmetry groups generated by at most three reflections. Each reflection is represented using the Hesse normal form~\cite{bocher1915plane, duda1972use}, enabling efficient and parallelizable symmetry search~\cite{je2024robust}. When multiple symmetries exist, we select the one corresponding to the fundamental domain with the smallest cardinality. In practice, if a symmetry group is generated by fewer than three reflections, we pad the remaining slots with a special symbol to maintain consistent data dimensionality.

Since each part is generated separately, symmetry is enforced at the part level. During part generation, we sample only the fundamental domain $\rv{d}$ and recover the full part by applying the learned symmetry group as defined in \cref{eq:symmetry-defn}. This approach reduces the number of points to generate and, more importantly, guarantees symmetry in the resulting shapes.

% Given a set dataset $\mathcal{P} \subseteq \mathbb{R}^{3 \times N}$ of the same part across all point clouds,
% To identify the symmetry group of a point cloud, Mitra~\etal~\cite{} proposes to use the mean-shift clustering~\cite{fukunaga1975estimation, comaniciu2002mean},  While it has shown promising results in detecting symmetries \cite[\eg,][]{chang2015shapenet}, it may fail under noisy shapes or noisy transformation space~\cite{je2024robust}. 

To learn the distribution of symmetry groups present in the dataset, we first construct a metric space $(\transspace, d_\transspace)$ of reflectional symmetries following
% Secs. 4.2 and 4.3 in
Je~\etal~\cite{je2024robust}. We then obtain the ground truth for symmetry groups $\mathcal{S}$ using mean-shift clustering~\cite{mitra2006partial, fukunaga1975estimation, comaniciu2002mean}, a nonparametric method based on gradient ascent over a density function in $\transspace$ (see \cref{sec:preliminary-symmetry} for more details). Finally, we leverage a diffusion model~\cite{ho2020denoising} to learn $p\left(\mathcal{S}\right)$.
% We then establish the connection between the mean-shift clustering and the annealed stochastic gradient Langevin dynamics~\cite{welling2011bayesian, song2019generative}, where stochastic noise is injected to improve the sample quality and to prevent the mode collapse.

More specifically, we apply the diffusion process (\cref{eq:diffusion-process} in \cref{sec:preliminary-diffusion}) to $\mathcal{S}$, and note that each transition kernel $q(\mathcal{S}_t \mid \mathcal{S}_{t-1}) \sim \normaldist{ \mu_t \mathcal{S}_{t-1}}{\sigma_t I}$ is Gaussian. Therefore, intermediate samples $\mathcal{S}_t$ can be expressed in closed form as
\begin{align}
    \mathcal{S}_t =  \left( \prod_{i=1}^t \mu_t \right) \mathcal{S}_0 + \sqrt{\sum_{i=1}^t \sigma_i^2 \prod_{j=i+1}^t \mu_j^2} \rv{\epsilon},
\end{align}
\hspace{-0.5em}
where $\rv{\epsilon} \sim \standardnormaldist$.
Setting $\mu_t \coloneqq 1$ and
% $\gamma_t \coloneqq \sqrt{\sum_{i=1}^t \sigma_i^2 \prod_{j=i+1}^t \mu_j^2}$ 
$\gamma_t \coloneqq \sqrt{\sum_{i=1}^t \sigma_i^2}$ 
simplifies this to
$
    \mathcal{S}_t =  \mathcal{S}_0 + \gamma_t \rv{\epsilon}.
$
Thus, the distribution of $\mathcal{S}_t$ becomes a convolution:
\begin{equation}
    p\left(\mathcal{S}_t\right) = \left( p_0 * \normalpdf{\rv{0}}{\gamma_t} \right) \left( \mathcal{S}_t \right) = \int p_0\left(\mathcal{S}\right) \normalpdf{\rv{0}}{\gamma_t}\left(\mathcal{S}_t-\mathcal{S}\right) \dd \mathcal{S} = \int p_0\left(S\right) \normalpdf{\mathcal{S}}{\gamma_t}\left(\mathcal{S}_t\right) \dd \mathcal{S},
\end{equation}
where $p_0$ is the distribution of $\mathcal{S}_0$, $\normalpdf{\mathcal{S}}{\gamma_t}$ is the probability density function for $\normaldist{\mathcal{S}}{\gamma_t I}$, and $*$ denotes convolution. With this, we train a diffusion model~\cite{ho2020denoising} $\symmetrydiffusion$ parametrized by $\rv{\zeta}$ to approximate the time-dependent score function~\cite{song2020score} 
%~\cite{song2020score, song2019generative}
(\ie the gradient of the log-density of noisy data) via an empirical approximation~\cite{je2024robust}:
%~\cite{je2024robust, karras2022elucidating, song2020score, pabbaraju2023provable}:
\begin{equation}
\label{sq:score-defn}
    \symmetrydiffusion \approx \nabla_{\mathcal{S}} \log p(\mathcal{S}_t) \approx \frac{1}{\gamma_t^2} \left( \frac{\sum_{R\in\transspace}\normalpdf{R}{\gamma_t}\left(\mathcal{S}\right)R}{\sum_{R\in\transspace}\normalpdf{R}{\gamma_t}\left(\mathcal{S}\right)} - \mathcal{S} \right).
\end{equation}
Since symmetry can vary across shapes, in practice, we condition the diffusion model on the shape latent $\prior$, modeling the symmetry group distribution as $p_\rv{\zeta}\left( \mathcal{S} \mid \prior \right)$.

After training, symmetries can be sampled from $\symmetrydiffusion$ through annealed stochastic gradient Langevin dynamics~\cite{song2019generative}: 
%~\cite{welling2011bayesian, song2019generative}
We initialize $\mathcal{S}_\tau^{(0)} \sim \standardnormaldist$
% from a prior distribution $p\left(\mathcal{S}_\tau^{(0)}\right) \sim \standardnormaldist$
and sequentially sample from noise-perturbed distributions $\symmetrydiffusion$ for $t=\tau, \tau-1, \ldots, 1$ using $L$ Langevin steps:
% where the initial sample $\mathcal{S}_t^{(i+1)}$ of each subsequent distribution is the final sample $\mathcal{S}_{t+1}^{(L)}$ of the previous one:
\begin{align}
    \mathcal{S}_t^{(0)} &\leftarrow \left\{\begin{array}{cl}
  \rv{\epsilon}_\tau^{(0)} &\text{if } t = \tau \\
 \mathcal{S}_{t+1}^{(L)} &\text{otherwise} \\ 
\end{array}\right.;\\
    \mathcal{S}_t^{(i+1)} &\leftarrow \mathcal{S}_t^{(i)} + \beta_t s_\rv{\zeta}\left( \mathcal{S}_t^{(i)} , t \right) + \sqrt{2\beta_t} \rv{\epsilon}_t^{(i)}, \quad i=0,1,\ldots,L, \label{eq:langevin-dynamics}
\end{align}
where $\rv{\epsilon}_t^{(i)} \sim \standardnormaldist$, and $\beta_t$ is the step size. The final sample $\mathcal{S}_1^{(L)}$ is the generated symmetry.

Notably, substituting \cref{sq:score-defn} into \cref{eq:langevin-dynamics} and setting $\beta_t \coloneqq \gamma_t^2$ yields the following update rule:
{\small
\begin{equation}
\label{eq:langevin-dynamics-conocrete}
    \mathcal{S}_t^{(i+1)} \leftarrow \mathcal{S}_t^{(i)} + \frac{\beta_t}{\gamma_t^2} \left( \frac{\sum_{T\in\transspace}\normalpdf{T}{\gamma_t}\left(\mathcal{S}_t^{(i)}\right)T}{\sum_{T\in\transspace}\normalpdf{T}{\gamma_t}\left(\mathcal{S}_t^{(i)}\right)} - \mathcal{S}_t^{(i)} \right) + \sqrt{2\beta_t} \rv{\epsilon} \overset{\beta_t \coloneqq \gamma_t^2}{=}
     \frac{\sum_{T\in\transspace}\normalpdf{T}{\gamma_t}\left(\mathcal{S}_t^{(i)}\right)T}{\sum_{T\in\transspace}\normalpdf{T}{\gamma_t}\left(\mathcal{S}_t^{(i)}\right)}+ \sqrt{2} \gamma_t \rv{\epsilon}.
\end{equation}
}
\hspace{-0.5em}
This update rule resembles mean-shift clustering (\cref{eq:mean-shift-iteration} in \cref{sec:preliminary-symmetry}), except it assumes an infinite neighborhood $B(\mathcal{S}_t) = \transspace$ and uses a Gaussian kernel with bandwidth $\gamma_t$:
\begin{equation}
    K_t(\mathcal{S}) \coloneqq \frac{1}{\sqrt{2 \pi} \gamma_t}e^{-\frac{\norm{\mathcal{S}}^2}{2}}.
\end{equation}
Another difference is the injected noise $\sqrt{2} \gamma_t \rv{\epsilon}$ in \cref{eq:langevin-dynamics-conocrete}, which has been shown to improve robustness, sample quality, and mitigate mode collapse~\cite[\eg][]{je2024robust}.
%~\cite{je2024robust, song2019generative, jolicoeur2020adversarial, song2020improved}.

\Cref{fig:show_symmetry} shows examples of identified symmetry group generators. Most parts exhibit symmetry of a single reflection. In addition, we note that classical symmetries, such as reflectional and rotational, are special cases of our formulation, where the symmetry group $\mathcal{S}$ is generated by a single transformation. Our approach generalizes this by allowing symmetries composed of multiple transformations, such as sequential reflections (\eg, chair legs in (e)) or a rotation followed by a reflection (\eg, car wheels). While our model does not explicitly capture circular symmetry, it can be approximated using discrete rotations with a minimum angle of $\frac{\pi}{18}$ (\eg, chair seat in (f)).

\subsection{Part Generation}
\label{sec:method-part-generation}
Given the identified symmetry groups $\symmetryj$ and corresponding fundamental domains $\fdj$ for each part $\partj$, we model the part distribution via the distribution of fundamental domains. Since each full part $\partj$ can be reconstructed by applying $\symmetryj$ to $\fdj$ (\cref{eq:symmetry-defn}), we have
\begin{equation}
  \partdist = \fddiffusion.
\end{equation}

To learn $\fddiffusion$, we employ a transformer-based diffusion model~\cite{mo2023dit, peebles2023scalable, vaswani2017attention}, conditioned on the shape latent $\prior$ and symmetry group $\symmetryj$. The model operates directly on voxelized point clouds, using 3D positional and patch embeddings to capture spatial context. Unlike U-Net architectures~\cite{ho2020denoising}, which face limitations in scalability and spatial coherence, our transformer backbone incorporates 3D window attention, reducing computational cost while preserving local structure. This design yields higher efficiency, scalability, and fidelity in point cloud generation~\cite{mo2023dit}.

% After training, to generate a part $\partj$, we first sample a shape latent $\rv{z}$ from $\priordiffusion$ and a symmetry group from $p_\rv{\zeta}\left( \mathcal{S}_j \mid \prior \right)$ learned in the previous sections. We then sample a fundamental domain $\fdj$ from the part diffusion $\fddiffusion$. Finally, we apply the identified symmetry group $\symmetryj$ on $\fdj$ to recover the full part $\partj$ as described in \cref{eq:symmetry-defn}. 
% As a result, we have
% \begin{equation}
%     p_\rv{\xi}(\partj, \symmetryj, \prior) = \fddiffusion p_\rv{\zeta}\left( \mathcal{S}_j \mid \prior \right) \priordiffusion.
% \end{equation}

\subsection{Part Assembly}
\label{sec:method-part-assembly}
To assemble the full point cloud, we learn assemblers $\transj$ that correctly transform each part $\partj$. We first train a part encoder $\partenc$ to obtain compact representations $\partembj$, then train a diffusion model $\assemblerdist$ to model the assembler distribution $\assemblerdist$. To simplify learning, we assume that $\partembj$ captures all necessary information about $\partj$ and $\symmetryj$, making $\transj$ conditionally independent of them given $\partembj$. Consequently, the diffusion model is conditioned only on $\partembj$ and the global shape latent $\prior$:
\begin{equation}
    \assemblerdiffusion \approx \assemblerdist.
\end{equation}
This assumption decouples geometric modeling from transformation learning, reducing complexity and allowing the model to focus on learning flexible, generalizable spatial configurations.

Our assembler diffusion model~\cite{tang2024diffuscene} employs a U-Net backbone~\cite{ronneberger2015u} augmented with skip connections and cross-attention layers~\cite{vaswani2017attention}. The cross-attention modules integrate contextual information from both part latents and the global shape latent, enabling the model to generate transformations that are coherent at both local and global levels. This architecture effectively handles diverse part configurations and generalizes well to novel shapes. The hierarchical structure of U-Net further supports multi-scale spatial reasoning, which is essential for accurate 3D part placement.

For the part encoder, we adopt the SVAE architecture (see \cref{sec:prior-distribution}) with \emph{equivariance fine-tuning} (\emph{EFT}): During training, random rigid transformations, including translations, rotations, and reflections, are applied to each part, and the encoder is encouraged to produce the latent transformed accordingly. This promotes geometry-aware but pose-invariant part embeddings, enhancing their suitability for the part assembly task. As a result, the diffusion model receives more stable and semantically meaningful latents, leading to more accurate and coherent assembler predictions.
\begin{table}
\centering
\caption{Quantitative comparison of point cloud generation. PA denotes part awareness; SA denotes symmetry awareness. Our \ours is the only model that supports both, achieving significant improvements over most baselines and setting a new state of the art.}
\label{tab:exp-table-big}
\setlength\tabcolsep{3pt}
\resizebox{\linewidth}{!}{
\begin{tabular}{lcccccccccccccc}
\toprule
\multicolumn{1}{c}{\multirow{3}{*}{Model}} & \multicolumn{1}{c}{\multirow{3}{*}{PA}} & \multicolumn{1}{c}{\multirow{3}{*}{SA}} & \multicolumn{4}{c}{Airplane} & \multicolumn{4}{c}{Car} & \multicolumn{4}{c}{Chair} \\
\cmidrule(lr){4-7} \cmidrule(lr){8-11} \cmidrule(lr){12-15}
& & & \multicolumn{2}{c}{1-NNA ($\downarrow$)} & \multicolumn{2}{c}{SDI ($\downarrow$)} & \multicolumn{2}{c}{1-NNA ($\downarrow$)} & \multicolumn{2}{c}{SDI ($\downarrow$)} & \multicolumn{2}{c}{1-NNA ($\downarrow$)} & \multicolumn{2}{c}{SDI ($\downarrow$)} \\
& & & CD & EMD & CD & EMD & CD & EMD & CD & EMD & CD & EMD & CD & EMD \\
\midrule
Training set & & & 64.4 & 64.1 & 0.954 & 4.90 & 51.3 & 54.8 & 7.49 & 1.18 & 51.7 & 50.0 & 5.56 & 1.68 \\
\cmidrule{1-15}
 % r-GAN~\cite{achlioptas2018learning} & \xmark & \xmark & 98.4 & 96.8 & 4519 & 1410 & 83.7 & 99.7 & 1053 & 362 & 94.5 & 99.0 & 7249 & 619 \\
 % 1-GAN/CD~\cite{achlioptas2018learning} & \xmark & \xmark & 87.3 & 94.0 & 3629 & 1353 & 68.6 & 83.8 & 918 & 352 & 66.5 & 88.8 & 7972 & 582 \\
 % 1-GAN/EMD~\cite{achlioptas2018learning} & \xmark & \xmark & 89.5 & 76.9 & 4129 & 914 & 71.9 & 64.7 & 982 & 313 & 71.2 & 66.2 & 7184 & 521 \\
 PointFlow~\cite{yang2019pointflow} & \xmark & \xmark & 75.7 & 70.7 & 3410 & 782 & 62.8 & 60.6 & 679 & 347 & 58.1 & 56.3 & 7290 & 530 \\
 % SoftFlow~\cite{kim2020softflow} & \xmark & \xmark & 76.1 & 65.8 & 3284 & 529 & 59.2 & 60.1 & 1549 & 428 & 64.8 & 60.1 & 5628 & 420 \\
 ShapeGF~\cite{cai2020learning} & \xmark & \xmark & 81.2 & 80.9 & 332 & 98.6 & 58.0 & 61.3 & 645 & 40.9 & 61.8 & 57.2 & 1100 & 101 \\
 DPF-Net~\cite{klokov2020discrete} & \xmark & \xmark & 75.2 & 65.6 & 4256 & 245 & 62.0 & 58.5 & 827 & 452 & 62.4 & 54.5 & 5234 & 245 \\
 SetVAE~\cite{kim2021setvae} & \xmark & \xmark & 76.5 & 67.7 & 2830 & 824 & 58.8 & 60.6 & 1240 & 327 & 59.9 & 59.9 & 5320 & 673 \\
 DPC~\cite{luo2021diffusion} & \xmark & \xmark & 76.4 & 86.9 & 187 & 44.2 & 60.1 & 74.8 & 217 & 30.3 & 68.9 & 80.0 & 335 & 50.6 \\
 PVD~\cite{zhou20213d} & \xmark & \xmark & 73.8 & 64.8 & 150 & 42.0 & 56.3 & 53.3 & 213 & 31.2 & 54.6 & 53.8 & 275 & 58.4 \\
 LION~\cite{vahdat2022lion} & \xmark & \xmark & 67.4 & 61.2 & 97.2 & 40.6 & 53.7 & 52.3 & 168 & 30.8 & 53.4 & 51.1 & 201 & 55.2 \\
 SPAGHETTI~\cite{hertz2022spaghetti} & \cmark & \xmark & 78.2 & 77.0 & 1530 & 529 & 72.3 & 71.0 & 581 & 284 & 70.7 & 69.0 & 5930 & 582 \\
 DiT-3D~\cite{mo2023dit} & \xmark & \xmark & 64.7 & 60.3 & 105 & 42.4 & 52.7 & \textbf{50.2} & 206 & 327 & 52.5 & 53.1 & 235 & 49.0 \\
  SALAD~\cite{koo2023salad} & \cmark & \xmark & 73.9 & 71.1 & 198 & 45.1 & 59.2 & 57.2 & 236 & 29.4 & 57.8 & 58.4 & 308 & 52.6 \\
 FrePolad~\cite{zhou2024frepolad} & \xmark & \xmark & 65.3 & 62.1 & 94.1 & 38.1 & 52.4 & 53.2 & 173 & 29.6 & 51.9 & \textbf{50.3} & 252 & 50.9 \\
\ours (ours) & \cmark & \cmark & \textbf{63.3} & \textbf{59.7} & \textbf{25.7} & \textbf{1.87} & \textbf{50.1} & 51.8 & \textbf{25.7} & \textbf{2.28} & \textbf{51.6} & 53.7 & \textbf{28.9} & \textbf{2.86} \\ 
 \bottomrule
\end{tabular}
}
\end{table}

\section{Experiment}
\label{sec:experiments}
\subsection{Dataset}
We evaluate our method on the ShapeNetPart dataset~\cite{yi2016scalable}, a subset of ShapeNet~\cite{chang2015shapenet} with semantic part annotations. We focus on three representative categories: airplanes (body, tail, wings, engines), cars (sunroof, wheels, hood, body), and chairs (backrest and armrests, legs, seat). Segmentation examples are shown in \cref{fig:show_symmetry}. We use the official train/validation/test split provided with the dataset.

Following common practice, we use point clouds with 2048 points. Each part is resized to match the mean number of points per part within its category.
% Since the number of points per part varies across instances, we apply a preprocess step to standardize part sizes for training. Specifically, for each part, we resample its point cloud to match the mean number of points for that part observed across the dataset.
During preprocessing, parts with more points are randomly subsampled, while those with fewer are upsampled via random duplication. This normalization ensures consistent input dimensions across training batches, enhancing the stability and efficiency of both the encoder and generative models.
% While simple, this resampling strategy preserves the geometric structure of each part while enabling fair comparisons across categories and instances.

\subsection{Evaluation Metrics}
Following prior work~\cite[\eg][]{zhou2024frepolad},
% ~\cite{vahdat2022lion, luo2021diffusion, zhou20213d, yang2019pointflow, zhou2024frepolad},
we assess the quality and diversity of generated point clouds using 1-nearest neighbor (1-NNA)~\cite{lopez2016revisiting}, a retrieval-based metric computed with Chamfer Distance (CD) and Earth Mover’s Distance (EMD). 1-NNA measures how often a shape’s nearest neighbor belongs to the same distribution~\cite{yang2019pointflow}. A balanced score near 50\% indicates close alignment between generated and real shape distributions.
% \Cref{sec:supp-full-table} in the supplementary provides further comparisons with two other metrics: minimum matching distance (MMD)~\cite{achlioptas2018learning} and coverage (COV)~\cite{achlioptas2018learning}.

To assess whether a generated 3D shape is symmetric,
% we manually select the most representative symmetry group $\mathcal{S}$ for each part based on the dataset (shown in \cref{fig:show_symmetry}; for chair, we choose the one for chair (e)). 
we introduce the \emph{symmetry discrepancy index (SDI)}, which quantifies how well a 3D shape $\rv{p}$ conforms to a given symmetry group $\mathcal{S}$: For a normalized shape $\rv{p}$, SDI is defined as the distance $d$ --- either CD or EMD --- between $\rv{p}$ and the shape reconstructed from its fundamental domain $\rv{d}$ under $\mathcal{S}$:
\begin{equation}
    \mathcal{L}_\text{SDI}(\rv{p}) \coloneqq d\left(\rv{p}, \mathcal{S} \rv{d} \right).
\end{equation}
Lower SDI values indicate stronger symmetry.
For methods without explicit symmetry modeling, we compute SDI using the simplest yet most common symmetry: reflection across the vertical bisector plane. For readability, we report SDI-CD scaled by $10$ and SDI-EMD scaled by $10^3$.
% For the SDI in \cref{tab:exp-table-per-part} where we do have access  methods supporting part awareness and compute SDI for the symmetry group shown in \cref{fig:show_symmetry} (for chair we choose the symmetry group of chair (e)).
% Note that this metric requires access to the fundamental domain $\rv{d}$ for the given $\mathcal{S}$. For point clouds generated by our \ours, $\rv{d}$ is automatically identified through the model; For other baselines, we manually extract $\rv{d}$ to enable a fair comparison.

\subsection{Point Cloud Generation}
\label{sec:experiment-Point Cloud Generation}
\begin{figure}
    \centering
    \includegraphics[width=\linewidth]{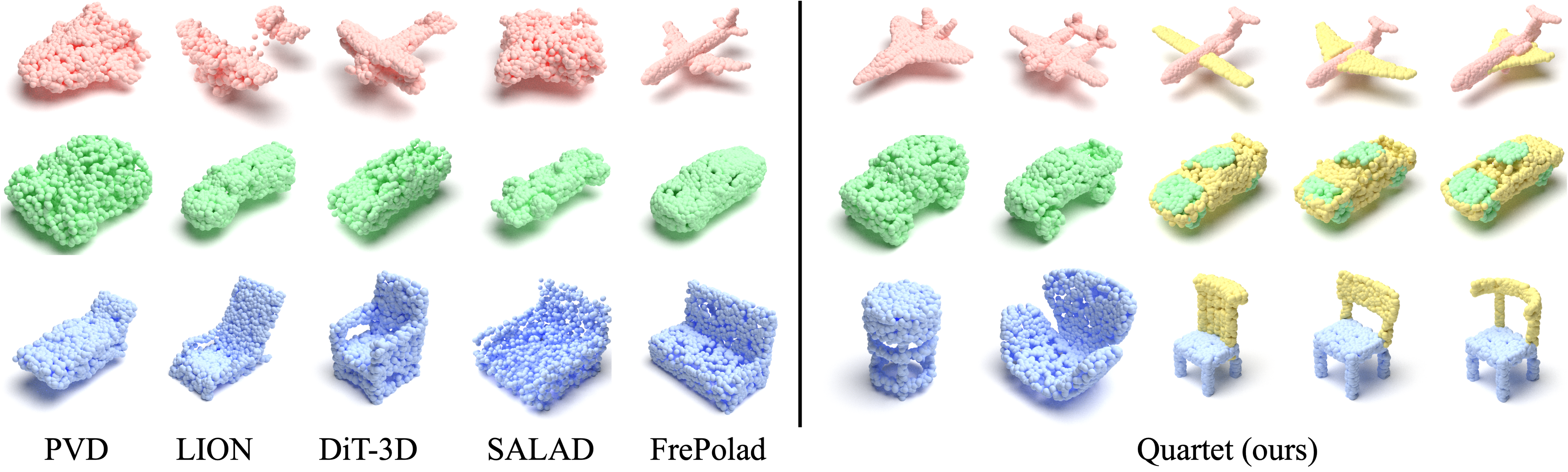}
    \caption{Point cloud generation. Samples from the \ours are visually appealing, diverse, and exhibit strong structural consistency. The last three columns illustrate targeted manipulation.}
    \label{fig:exp-gen}
\end{figure}

We benchmark the \ours against several competitive 3D generative models. Implementation and training details are provided in \cref{sec:Training Details}, and additional experimental results can be found in \cref{sec:appendix-full-experiments-results}. Quantitative results based on 1-NNA and SDI are reported in \cref{tab:exp-table-big}. Notably, the \ours is the only method that explicitly models both part structure and symmetry. Per-part SDI scores for part-aware models are presented in \cref{tab:exp-table-per-part}. The \ours consistently outperforms state-of-the-art methods, achieving high fidelity and diversity while closely matching the real shape distribution. Remarkably, due to explicit symmetry enforcement, the \ours achieves significantly lower SDI scores, indicating strong alignment with ideal symmetric structures. 
Qualitative results in \cref{fig:exp-gen} show that the \ours generates visually coherent and diverse point clouds obeying symmetry across all three object categories, demonstrating its effectiveness.

\paragraph{Targeted manipulation} The \ours models the distribution of each part separately, enabling targeted part-level manipulation while guaranteeing disentanglement. In \cref{fig:teaser} and in the last three columns of \cref{fig:exp-gen}, we present point cloud samples in which the yellow-highlighted parts vary while the remaining structure is held fixed. 
\begin{table}
\centering
\caption{Per-part SDI-CD ($\downarrow$) for point cloud generation. With explicit symmetry enforcement, the \ours achieves significantly lower SDI scores, indicating stronger symmetry in generated parts.}
\label{tab:exp-table-per-part}
\setlength\tabcolsep{4.7pt}
\resizebox{\linewidth}{!}{
\begin{tabular}{lccccccccccc}
\toprule
\multicolumn{1}{c}{\multirow{2}{*}{Model}} & \multicolumn{4}{c}{Airplane} & \multicolumn{4}{c}{Car} & \multicolumn{3}{c}{Chair} \\
\cmidrule(lr){2-5} \cmidrule(lr){6-9} \cmidrule(lr){10-12}
  & body & tail & wings & engines & roof & wheels & hood & body & back & legs & seat \\
\midrule
Training set & 0.877 & 0.700 & 0.883 & 0.717 & 5.25 & 3.06 & 3.15 & 5.27 & 4.11 & 2.65 & 3.04 \\
\cmidrule{1-12}
 SPAGHETTI~\cite{hertz2022spaghetti} & 318 & 268 & 417 & 125 & 258 & 339 & 127 & 417 & 2857 & 2948 & 1358 \\
  SALAD~\cite{koo2023salad} & 39.1 & 26.2 & 30.5 & 18.5 & 98.2 & 68.5 & 152 & 256 & 163 & 93.5 & 104 \\
\ours (ours) & \textbf{7.86} & \textbf{4.08} & \textbf{7.76} & \textbf{5.84} & \textbf{9.72} & \textbf{4.10} & \textbf{5.76} & \textbf{9.90} & \textbf{10.3} & \textbf{6.19} & \textbf{10.4} \\ 
 \bottomrule
\end{tabular}
}
\end{table}

\subsection{Ablation Study}
\label{sec:experiment-ablation}
Our \ours consists of four diffusion models responsible for shape latents, symmetries, parts, and assemblers. As an ablation study, in \cref{tab:ablation}, we evaluate several simplified variants with some diffusions or key techniques removed:
\begin{itemize}
    \item NoSVAE: replaces SVAEs (\cref{sec:prior-distribution}) with standard VAEs for full point cloud and part encoding.
    \item NoEFT: removes the equivariance fine-tuning (\cref{sec:method-part-assembly}) applied to the part SVAE.
    \item Trio Var. 1: replaces shape latent diffusion with a simple Gaussian prior;
    \item Trio Var. 2: removes the symmetry diffusion with part diffusion generating the full part directly;
    \item Duet Var. 1: removes both shape latent and symmetry diffusions;
    \item Duet Var. 2: removes symmetry diffusion and merges part and assembler diffusions into a single full-shape diffusion;
    \item Solo: uses a single diffusion generating full point clouds.
\end{itemize}
Our results show that all four diffusion models are indispensable members of the \ours, each contributing to the generation of high-quality, diverse point clouds with strong symmetry. Two key observations emerge: First, comparing Solo and Duet Var. 1, we see that generating parts without a central shape latent significantly degrades performance. This echoes previous findings that a simple Gaussian prior is insufficient for capturing the structural complexity of 3D shapes~\cite{vahdat2021score, tomczak2018vae, rosca2018distribution, zhou2024frepolad, vahdat2022lion}. The performance drops further in Trio Var. 1, where symmetry diffusion is added, likely compounding the mismatch between prior and part representations. Second, comparing Trio Var. 2 and the full \ours, we observe that while symmetry enforcement substantially improves SDI, it does not consistently improve 1-NNA -- even with latent diffusion present. This may be due to the added constraint or increased learning complexity introduced by enforcing symmetry.
\begin{table}
\centering
\caption{Ablation study. L denotes latent diffusion, S symmetry diffusion, and P part and assembler diffusions. All four members in the \ours are essential; removing any degrades performance.}
\label{tab:ablation}
\setlength\tabcolsep{3.5pt}
\resizebox{\linewidth}{!}{
\begin{tabular}{lccccccccccccccc}
\toprule
\multicolumn{1}{c}{\multirow{3}{*}{Variation}} & \multicolumn{1}{c}{\multirow{3}{*}{L}} & \multicolumn{1}{c}{\multirow{3}{*}{S}} & \multicolumn{1}{c}{\multirow{3}{*}{P}} & \multicolumn{4}{c}{Airplane} & \multicolumn{4}{c}{Car} & \multicolumn{4}{c}{Chair} \\
\cmidrule(lr){5-8} \cmidrule(lr){9-12} \cmidrule(lr){13-16}
& & & & \multicolumn{2}{c}{1-NNA ($\downarrow$)} & \multicolumn{2}{c}{SDI ($\downarrow$)} & \multicolumn{2}{c}{1-NNA ($\downarrow$)} & \multicolumn{2}{c}{SDI ($\downarrow$)} & \multicolumn{2}{c}{1-NNA ($\downarrow$)} & \multicolumn{2}{c}{SDI ($\downarrow$)} \\
& & & & CD & EMD & CD & EMD & CD & EMD & CD & EMD & CD & EMD & CD & EMD \\
\midrule
Solo & \xmark & \xmark & \xmark & 69.2 & 64.2 & 154 & 44.2 & 55.5 & 53.6 & 241 & 351 & 53.2 & 53.8 & 295 & 52.8 \\ 
Duet Var. 1 & \xmark & \xmark & \cmark & 76.3 & 82.1 & 927 & 614 & 70.2 & 72.9 & 522 & 460 & 68.3 & 67.2 & 3261 & 615 \\
Duet Var. 2 & \cmark & \xmark & \xmark & 66.8 & 63.3 & 103 & 42.4 & 52.3 & 52.7 & 192 & 28.1 & 53.2 & 51.6 & 184 & 49.2 \\
Trio Var. 1 & \xmark & \cmark & \cmark & 85.1 & 85.8 & 3516 & 1450 & 83.9 & 83.6 & 836 & 326 & 92.5 & 85.2 & 6286 & 562 \\
Trio Var. 2 & \cmark & \xmark & \cmark & \textbf{63.1} & 63.6 & 95 & 39.2 & \textbf{49.8} & 52.4 & 205 & 32.3 & 52.3 & 53.9 & 174 & 52.5 \\
NoSVAE & \cmark & \cmark & \cmark & 64.3 & 61.8 & \textbf{25.2} & 2.51 & 51.6 & 52.0 & 25.9 & 2.55 & 52.6 & 53.9 & 29.3 & 3.15 \\
NoEFT & \cmark & \cmark & \cmark & 63.9 & 62.4 & 32.7 & 9.27 & 51.8 & 52.0 & 27.2 & 5.98 & 52.5 & 54.0 & 30.2 & 4.15 \\
\ours & \cmark & \cmark & \cmark & 63.3 & \textbf{59.7} & 25.7 & \textbf{1.87} & 50.1 & \textbf{51.8} & \textbf{25.7} & \textbf{2.28} & \textbf{51.6} & \textbf{53.7} & \textbf{28.9} & \textbf{2.86}  \\ 
 \bottomrule
\end{tabular}
}
\end{table}

\subsection{Runtime Analysis}
\label{sec:Runtime analysis}
Training the \ours on each object category takes approximately 50 hours on a single GPU. \Cref{fig:teaser} presents the generation performance \vs training time for models trained on the airplane category. Some baselines require longer training due to joint optimization~\cite{yang2019pointflow}, operation on full point clouds~\cite{zhou20213d, luo2021diffusion}, or the use of complex, high-dimensional latent spaces~\cite{vahdat2022lion}. In contrast, despite using four diffusions, the \ours trains faster than most baselines. This is because the shape latent, symmetry, and assembler diffusions operate on low-dimensional representations ($124$-, $12M$-, and $9M$-dimensional, respectively, where $M$ is the number of parts). The majority of training time is spent on part diffusion, which is accelerated using an efficient transformer-based diffusion~\cite{mo2023dit}
%~\cite{mo2023dit, peebles2023scalable, vaswani2017attention}
applied to fundamental domains that are typically only half or a quarter the size of the full parts.

\section{Conclusion and Discussion}
\label{sec:Conclusion and Discussion}
We presented the \oursfull, a structure-aware framework for 3D point cloud generation that explicitly models part composition and symmetry.
% By disentangling the generative process into four coordinated diffusion models, our method
% % enforces structural priors that are often neglected in existing approaches. This design
% enables an interpretable and controllable generation of diverse, high-quality point clouds with guaranteed symmetry and coherent part assembly.
% % The structured representation also makes the process interpretable and controllable, supporting fine-grained manipulation.
% We demonstrated the effectiveness of our method through extensive experiments and analyses. 
% At its current stage, this work focuses on unconditional generation using ShapeNet~\cite{chang2015shapenet}. Nonetheless, we hope it underscores the value of integrating symmetry- and part-based reasoning at the core of structure-aware generative models and opens new directions for future research. Promising extensions include adapting the \ours to more complex datasets or conditional generation tasks.
% % , such as those guided by natural language descriptions.
% Moreover, the model’s ability to disentangle and manipulate individual parts makes it well-suited for interactive shape editing and user-guided design applications.
At its current stage, our work focuses on unconditional generation,
% using the ShapeNet dataset,
but it highlights the importance of integrating symmetry- and part-based reasoning into structure-aware models.
% Future directions include extending the \ours to more complex datasets or conditional generation tasks.
% Moreover,
The model’s ability to disentangle and manipulate individual parts makes it well-suited for interactive shape editing and user-guided design applications.

\clearpage
\bibliography{main}
\bibliographystyle{iclr2026_conference}

%%%%%%%%%%%%%%%%%%%%%%%%%%%%%%%%%%%%%%%%%%%%%%%%%%%%%%%%%%%%

\clearpage
\appendix
\maketitlesupplementary

\section{Related Work on Symmetry Detection}
\label{sec:supp-symmetry-detection}
% Symmetry is a fundamental geometric property observed across natural and human-made objects. Therefore, detecting and leveraging symmetry has long been an important problem in computer vision and graphics~\cite{mitra2013symmetry}. Early work focuses on finding exact symmetries on 2D or 3D planar point sets~\cite{atallah1985symmetry, wolter1985optimal}, but the restriction to exact symmetries limits the use of these methods for real-world objects. Traditional voting-based methods work by aggregating votes on the symmetry via the Hough transform~\cite{illingworth1988survey} but struggle with noise. Mitra~\etal~\cite{mitra2006partial} improve this method by leveraging the mean-shift clustering~\cite{fukunaga1975estimation, comaniciu2002mean} in a transformation space (for more details see \cref{sec:preliminary-symmetry}). Their work is followed by further improvements~\cite{pauly2008discovering, shi2016symmetry}. More recently, learning based methods are also proposed~\cite{ji2019fast, zhou2021nerd, gao2020prs}. Je~\etal~\cite{je2024robust} combines the voting- and learning-based detection by applying Langevin dynamics to a redefined symmetry space for a robust symmetry detection.
Symmetry is a fundamental geometric property commonly observed in both natural and human-made objects. As a result, detecting and leveraging symmetry has long been a central problem in the fields of computer vision and computer graphics~\cite{mitra2013symmetry}. Early work in this area focused on detecting exact symmetries in 2D or 3D planar point sets~\cite{atallah1985symmetry, wolter1985optimal}. However, the reliance on exact symmetry limits the practicality of these methods, as real-world objects often exhibit only approximate symmetry due to noise, occlusion, or design variations.

Traditional voting-based approaches, such as those based on the Hough transform~\cite{illingworth1988survey}, attempt to accumulate votes for potential symmetries from point correspondences. While effective in idealized settings, these methods are known to be sensitive to noise and often produce unreliable results when applied to imperfect or incomplete data.

To address these limitations, Mitra~\etal~\cite{mitra2006partial} proposed a more robust technique that replaces voting with mean-shift clustering~\cite{fukunaga1975estimation, comaniciu2002mean} in a transformation space (for more details see \cref{sec:preliminary-symmetry}). This method searches for modes in the space of rigid transformations, making it more resilient to noise and better suited for partial symmetries. Their approach laid the groundwork for several subsequent improvements~\cite{pauly2008discovering, shi2016symmetry}, which further enhanced the robustness and generality of symmetry detection.

In recent years, learning-based methods have also been introduced~\cite{ji2019fast, zhou2021nerd, gao2020prs}, leveraging neural networks to detect symmetry directly from 3D data. These methods benefit from data-driven representations and have shown improved generalization across object categories and varying conditions.

Building on these trends, Je~\etal~\cite{je2024robust} proposed a hybrid approach that combines elements of both traditional and learning-based techniques. Their method redefines the symmetry space and applies Langevin dynamics—a sampling technique from generative modeling—to iteratively refine symmetry estimates. This formulation provides both robustness and efficiency and serves as a strong foundation for the symmetry detection component in our framework.

\section{Background}
\subsection{Symmetry Detection by Mean-Shift Clustering}
\label{sec:preliminary-symmetry}
A common approach to detecting symmetry in 3D objects involves using the Hough transform~\cite{illingworth1988survey} to vote on the parameters of potential symmetry planes~\cite{mitra2013symmetry, mitra2006partial}. In this framework, each pair of points in the shape casts a vote for a candidate symmetry in a predefined transformation space $\transspace$. For example, in the case of reflectional symmetry, a point pair $\rv{a}$ and $\rv{b}$ casts a vote for the plane that passes through their midpoint $\frac{\rv{a}+\rv{b}}{2}$ and has a normal vector given by $\frac{\rv{a}-\rv{b}}{\norm{\rv{a}-\rv{b}}}$ -- the direction that would reflect one point onto the other.

In an ideal setting, where the object and its symmetries are exact, votes from all symmetric point pairs would concentrate on discrete points in $\transspace$. In such cases, detecting symmetry reduces to identifying the peak with the highest vote count. However, most real-world objects and 3D datasets only exhibit approximate symmetry. As a result, the votes form a smooth, continuous distribution in $\transspace$ rather than sharp peaks. Detecting symmetry in this context thus requires identifying clusters in $\transspace$ that correspond to the most prominent approximate symmetries.

Mitra~\etal~\cite{mitra2006partial} proposes to use the mean-shift clustering~\cite{fukunaga1975estimation, comaniciu2002mean}, a nonparametric clustering method based on gradient ascent on a density function $p(S)$ in $\transspace$ defined as
\begin{equation}
\label{eq:mean-shift-density}
    p(S) \coloneqq \sum_{R \in \transspace} K\left(\frac{S-R}{h}\right),
\end{equation}
where $K$ is a kernel function (\eg, Gaussian or Epanechnikov kernel~\cite{epanechnikov1969non}) with bandwidth $h$. The significant modes of $p$, and hence the significant symmetries, can be determined using gradient ascent: The algorithm first initializes from any of the candidate transformation $S_0 \in \transspace$ and performs the following iteration until convergence:
\begin{equation}
\label{eq:mean-shift-iteration}
    S_{t+1} \leftarrow \frac{\sum_{R \in B(S_t)}  K\left(\frac{S_t-R}{h}\right) R}{\sum_{R \in B(S_t)}  K\left(\frac{S_t-R}{h}\right)},
\end{equation}
where $B(S_t)$ is a neighborhood of $S_t$.

While it has shown promising results in detecting symmetries \cite[\eg,][]{chang2015shapenet}, it may fail under noisy shapes or noisy transformation space~\cite{je2024robust}. Inspired by Je~\etal~\cite{je2024robust}, we leverage a diffusion process~\cite{ho2020denoising} to establish the connection between the iteration in \cref{eq:mean-shift-iteration} and the stochastic gradient Langevin dynamics~\cite{welling2011bayesian}, where stochastic noise is injected to improve the sample quality and robustness.

\subsection{Variational Autoencoder}
\label{sec:preliminary-vae}
We use \emph{variational autoencoders (VAEs)}~\cite{kingma2013auto} as our latent distribution model as it provides access to a low-dimensional latent space and has been successfully applied to generate point clouds~\cite{li2022editvae, wang2020learning, li2022spa, zhou2024frepolad}. VAEs are probabilistic generative models that can model a probability distribution of a given dataset $\pcdataset$.
% A VAE consists of an encoder $\pcenc$ and a decoder $\pcdec$ parametrized by $\rv{\eta}$.

Starting with a known prior distribution $\vaeprior$ of shape latents $\prior \in \mathbb{R}^\prior$, the parametric decoder of a VAE models the conditional distribution $\pcdec$ parametrized by $\rv{\eta}$. However, training the decoder to maximize the likelihood of data is not possible as
\begin{equation}
p(\pc) = \int \pcdec p(\rv{z}) \, \dd \rv{z}
\end{equation}
is intractable. Instead, a parametric encoder $\pcdec$ is used to approximates the posterior distribution. Both networks are jointly trained to maximize a lower bound on the likelihood called the \emph{evidence lower bound (ELBO)}:
\begin{equation}
    \begin{aligned}
\label{eq:vae-elbo-no-beta}
    \mathcal{L}_\text{ELBO} (\rv{\eta}; \pc) \coloneqq \mathbb{E}_{\pcenc} \left[ \log \pcdec \right] - \mathcal{D}_\text{KL} \left( \pcenc, \vaeprior \right),
\end{aligned}
\end{equation}
where $\mathcal{D}_\text{KL}$ is the Kullback-Leibler divergence between the two distributions~\cite{csiszar1975divergence}.

\subsection{Denoising Diffusion Probabilistic Model}
\label{sec:preliminary-diffusion}
Our generative framework leverages four \emph{denoising diffusion probabilistic models (DDPMs \textnormal{or} diffusion models)}~\cite{sohl2015deep, ho2020denoising} to model distinct data distributions. Given a data sample $\rv{z} \sim p(\rv{z})$, diffusion models progressively corrupt $\rv{z} = \rv{z}_0$ into a noisy sample $\rv{z}_\tau$ through a Markovian \emph{forward diffusion process}. At each time step $t = 1, 2, \ldots, \tau$, Gaussian noise is added according to a predefined variance schedule $\{\sigma_t\}_t$:
\begin{align}
q \left( \rv{z}_{1:\tau} \mid \rv{z}_0 \right) &:= \prod_{t=1}^\tau q ( \rv{z}_t \vert \rv{z}_{t-1} ) \\
q(\rv{z}_t \mid \rv{z}_{t-1}) &\sim \normaldist{ \mu_t \rv{z}_{t-1}}{\sigma_t I}, \label{eq:diffusion-process}
\end{align}
where $\normaldist{\rv{\mu}}{\rv{\sigma}}$ denotes multivariate Gaussian distribution with mean $\rv{\mu}$ and variance $\rv{\sigma}$. In practice, we set $\mu_t \coloneqq \sqrt{1-\sigma_t}$. If $\tau$ is sufficiently large (\eg, 1000 steps), $p(\rv{z}_\tau)$ will approach the standard Gaussian distribution $\normaldist{0}{I}$. 

Diffusion models learn a \emph{reverse process} $p_\rv{\theta}(\rv{z}_{t-1} \mid \rv{z}_t)$, parameterized by $\rv{\theta}$, which defines a Markovian denoising chain that inverts the forward diffusion. This process gradually transforms a sample of standard Gaussian noise $\rv{z}_\tau$ back into a data sample $\rv{z}_0$:
\begin{align}
\label{eq:markov}
p_\rv{\theta}(\rv{z}_{0:\tau}) &:= p(\rv{z}_\tau) \prod_{t=1}^\tau p_\rv{\theta}(\rv{z}_{t-1} \mid \rv{z}_t);\\
\label{eq:p-zt-1-given-zt}
p_\rv{\theta}(\rv{z}_{t-1} \mid \rv{z}_t) &\sim \normaldist{\rv{\mu}_\rv{\theta}(\rv{z}_t, t)}{\varsigma_t^2 I},
\end{align}
where $\rv{\mu}_\rv{\theta} (\rv{z}_t, t)$ represents the predicted mean for the Gaussian distribution at time step $t$ and $\{\varsigma_t\}_t$ is another variance schedule.

DDPMs are trained by maximizing the variational lower bound of log-likelihood of the data $\rv{z}_0$ under $q(\rv{z}_0)$:
\begin{align}
    \label{eq:variational-lower-bound}
    \mathbb{E}_{q(\rv{z}_0)} \left[ \log p_\rv{\theta}(\rv{z}_0) \right] \geq \mathbb{E}_{q(\rv{z}_{0:\tau})} \left[ \log \frac{p_\rv{\theta}(\rv{z}_{0:\tau})}{q(\rv{z}_{1:\tau} \mid \rv{z}_0)} \right].
\end{align}
Expanding \cref{eq:variational-lower-bound} with \cref{eq:markov} and noticing that $p(\rv{z}_\tau)$ and $q(\rv{z}_{1:\tau} \mid \rv{z}_0)$ are constant with respect to $\rv{\theta}$, we obtain our objective function to maximize:
\begin{align}
\label{eq:pre-objective}
    \mathbb{E}_{q(\rv{z}_0),q(\rv{z}_{1:\tau} \mid \rv{z}_0)} \left[ \sum_{t=1}^\tau \log p_\rv{\theta} ( \rv{z}_{t-1} \mid \rv{z}_t) \right].
\end{align}
Since we can factor the joint posterior
\begin{equation}
q(\rv{z}_{1:\tau} \mid \rv{z}_0) = \prod_{t=1}^\tau q(\rv{z}_{t-1} \mid \rv{z}_t, \rv{z}_0)
\end{equation}
and both $q(\rv{z}_{t-1} \mid \rv{z}_t, \rv{z}_0)$ and $p_\rv{\theta}(\rv{z}_{t-1} \mid \rv{z}_t)$ are Gaussian, maximizing \cref{eq:pre-objective} is equivalent as minimizing the following score matching objective for a parametric model $\rv{\epsilon}_\rv{\theta}(\rv{z}_t,t)$:
\begin{align}
    \label{eq:loss-diffusion}
    \mathcal{L}_\text{diffusion}(\rv{\theta}) \coloneqq \mathbb{E}_{p(\rv{z}_0),t \sim \mathcal{U}(1,\tau), \rv{\epsilon} \sim \normaldist{0}{I}} \left[ \| \rv{\epsilon} - \rv{\epsilon}_\rv{\theta}(\rv{z}_t,t) \|^2 \right],
\end{align}
where $\mathcal{U}(1,\tau)$ is the uniform distribution on $\{1,2,\ldots,\tau\}$. Intuitively, minimizing this loss corresponds to learning to predict the noise $\rv{\epsilon}$ required to denoise the diffused sample $\rv{z}_t$. Notably, the training objective of diffusion models closely aligns with estimating the gradient of the log data density -- \ie, the score function -- as used in score-based energy models~\cite{song2020denoising, swersky2011autoencoders, hyvarinen2009estimation, lecun2006tutorial}.

During inference, the network allows sampling through an iterative procedure since the learned distribution can be factorized as
\begin{equation}
\label{eq:diffusion-gen}
    p_\rv{\theta}\left( \left\{ \rv{z}_t \right\}_{t=0}^\tau\right) = p\left(\rv{z}_\tau \right) p_\rv{\theta}\left(\rv{z}_{t-1} \mid \rv{z}_t \right)  = p\left(\rv{z}_\tau \right) \prod_{t=1}^\tau p_\rv{\theta} \left(\rv{z}_{t-1} \mid \rv{z}_t \right)
\end{equation}
for $p(\rv{z}_\tau) \coloneqq \normaldist{0}{I}$.

\section{Implementation and Training Details}
\label{sec:Training Details}
% Technical appendices with additional results, figures, graphs and proofs may be submitted with the paper submission before the full submission deadline (see above), or as a separate PDF in the ZIP file below before the supplementary material deadline. There is no page limit for the technical appendices.
\Cref{tab:arch_details,tab:traing_details} provide the architecture and training details for the various VAE and diffusion models used in the \ours. The hyperparameter $\lambda$ in \cref{eq:svae-loss}, which controls the activation sparsity in the SVAE, is set to 0.005. The hyperparameter for the equivariance fine-tuning in the part VAE, discussed in \cref{sec:method-part-assembly}, is set to 0.01. For all models, we employ a learning rate scheduler with a reduce-on-plateau policy, which decreases the learning rate by a factor of 10 if the loss does not improve for 10 consecutive epochs. All training runs converged successfully by the end of training.
\begin{table}[ht]
    \centering
    \caption{Model architecture details for the \ours. $M$ denotes the number of parts for each object category ($M = 4$ for airplanes and cars; $M = 3$ for chairs).}
    \label{tab:arch_details}
    \resizebox{\linewidth}{!}{
    \begin{tabular}{lccc}
        \toprule
         Model & Backbone & Data dimensionality & Latent dimensionality \\
         \midrule
         Point cloud SVAE & PVCNN~\cite{liu2019point} & $2048 \times 3$ & $128$\\
         Shape latent diffusion & U-Net~\cite{ronneberger2015u} & $128$ & -\\
         Symmetry diffusion & ResNet~\cite{he2016deep} & $12 \times M$ & - \\
         Part diffusion & Transformer~\cite{vaswani2017attention} & $2048 \times 3$ & - \\
         Part VAE & PVCNN~\cite{liu2019point} & $57 \times 3$ to $1024 \times 3$ & $128$ \\
         Assembler diffusion & U-Net~\cite{ronneberger2015u} & $9 \times M$ & - \\
         \bottomrule
    \end{tabular}}
\end{table}
\begin{table}[ht]
    \centering
    \caption{Training details for the \ours.}
    \label{tab:traing_details}
    \begin{tabular}{lccc}
        \toprule
         Model & Batch size & Number of epoch & Learning rate \\
         \midrule
         Point cloud SVAE & $64$ & $1000$ & $10^{-3}$\\
         Shape latent diffusion & $32$ & $2000$ & $10^{-4}$\\
         Symmetry diffusion & $32$ & $2000$ & $10^{-4}$\\
         Part diffusion & $16$ & $2000$ & $10^{-3}$\\
         Part VAE & $64$ & $1000$ & $10^{-5}$ \\
         Assembler diffusion & $32$ & $2000$ & $10^{-4}$ \\
         \bottomrule
    \end{tabular}
\end{table}

\section{More Experimental Results}
\label{sec:appendix-full-experiments-results}

\begin{figure}
    \centering
    \includegraphics[width=\linewidth]{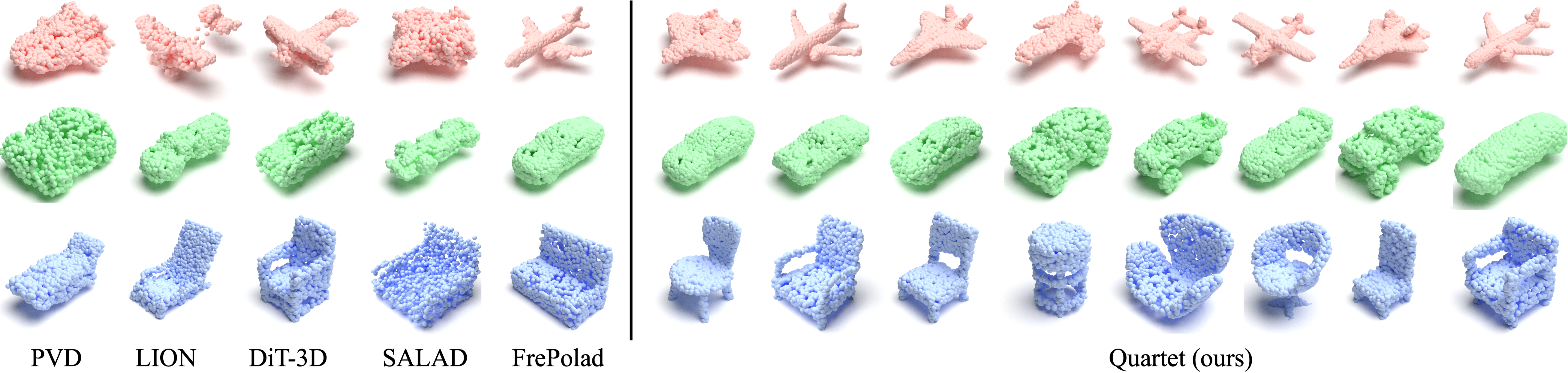}
    \caption{Full point cloud generation results. Samples from the \ours are visually appealing, diverse, and exhibit strong structural and symmetry consistency.}
    \label{fig:exp-gen-full}
\end{figure}

\begin{figure}
    \centering
    \includegraphics[width=\linewidth]{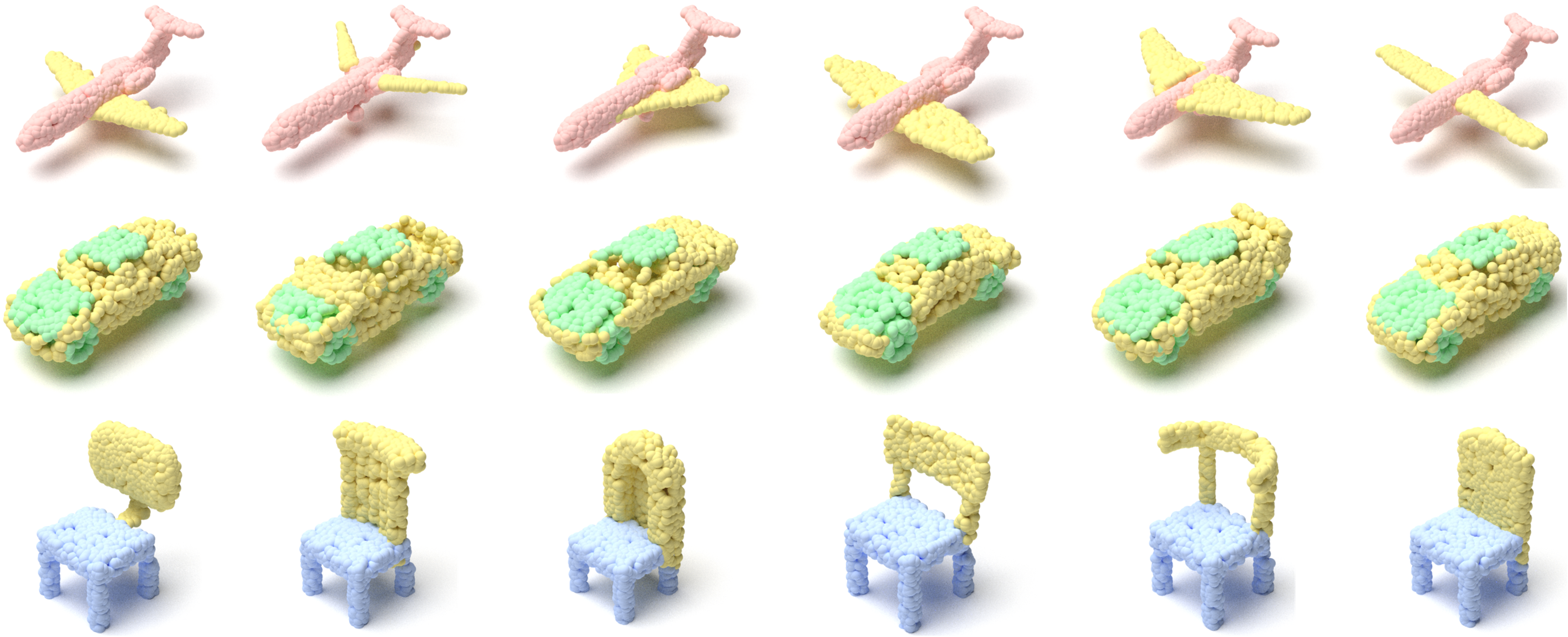}
    \caption{More targeted manipulation results. We vary the yellow-highlighted parts while holding the remaining structure fixed.}
    \label{fig:targeted-editing}
\end{figure}

\Cref{tab:exp-table-big-full,fig:exp-gen-full,fig:targeted-editing} present more results for the point cloud generation experiment in \cref{sec:experiment-Point Cloud Generation}.
\begin{table}
\centering
\caption{Quantitative comparison of point cloud generation. PA denotes part awareness; SA denotes symmetry awareness. Our \ours is the only model that supports both, achieving significant improvements over most baselines and setting a new state of the art.}
\label{tab:exp-table-big-full}
{
\setlength\tabcolsep{2pt}
\resizebox{\linewidth}{!}{
\begin{tabular}{lcccccccccccccc}
\toprule
\multicolumn{1}{c}{\multirow{3}{*}{Model}} & \multicolumn{1}{c}{\multirow{3}{*}{PA}} & \multicolumn{1}{c}{\multirow{3}{*}{SA}} & \multicolumn{4}{c}{Airplane} & \multicolumn{4}{c}{Car} & \multicolumn{4}{c}{Chair} \\
\cmidrule(lr){4-7} \cmidrule(lr){8-11} \cmidrule(lr){12-15}
& & & \multicolumn{2}{c}{1-NNA ($\downarrow$)} & \multicolumn{2}{c}{SDI ($\downarrow$)} & \multicolumn{2}{c}{1-NNA ($\downarrow$)} & \multicolumn{2}{c}{SDI ($\downarrow$)} & \multicolumn{2}{c}{1-NNA ($\downarrow$)} & \multicolumn{2}{c}{SDI ($\downarrow$)} \\
& & & CD & EMD & CD & EMD & CD & EMD & CD & EMD & CD & EMD & CD & EMD \\
\midrule
Training set & & & 64.4 & 64.1 & 0.954 & 4.90 & 51.3 & 54.8 & 7.49 & 1.18 & 51.7 & 50.0 & 5.56 & 1.68 \\
\cmidrule{1-15}
 r-GAN~\cite{achlioptas2018learning} & \xmark & \xmark & 98.4 & 96.8 & 4519 & 1410 & 83.7 & 99.7 & 1053 & 362 & 94.5 & 99.0 & 7249 & 619 \\
 1-GAN/CD~\cite{achlioptas2018learning} & \xmark & \xmark & 87.3 & 94.0 & 3629 & 1353 & 68.6 & 83.8 & 918 & 352 & 66.5 & 88.8 & 7972 & 582 \\
 1-GAN/EMD~\cite{achlioptas2018learning} & \xmark & \xmark & 89.5 & 76.9 & 4129 & 914 & 71.9 & 64.7 & 982 & 313 & 71.2 & 66.2 & 7184 & 521 \\
 PointFlow~\cite{yang2019pointflow} & \xmark & \xmark & 75.7 & 70.7 & 3410 & 782 & 62.8 & 60.6 & 679 & 347 & 58.1 & 56.3 & 7290 & 530 \\
 SoftFlow~\cite{kim2020softflow} & \xmark & \xmark & 76.1 & 65.8 & 3284 & 529 & 59.2 & 60.1 & 1549 & 428 & 64.8 & 60.1 & 5628 & 420 \\
 ShapeGF~\cite{cai2020learning} & \xmark & \xmark & 81.2 & 80.9 & 332 & 98.6 & 58.0 & 61.3 & 645 & 40.9 & 61.8 & 57.2 & 1100 & 101 \\
 DPF-Net~\cite{klokov2020discrete} & \xmark & \xmark & 75.2 & 65.6 & 4256 & 245 & 62.0 & 58.5 & 827 & 452 & 62.4 & 54.5 & 5234 & 245 \\
 SetVAE~\cite{kim2021setvae} & \xmark & \xmark & 76.5 & 67.7 & 2830 & 824 & 58.8 & 60.6 & 1240 & 327 & 59.9 & 59.9 & 5320 & 673 \\
 DPC~\cite{luo2021diffusion} & \xmark & \xmark & 76.4 & 86.9 & 187 & 44.2 & 60.1 & 74.8 & 217 & 30.3 & 68.9 & 80.0 & 335 & 50.6 \\
 PVD~\cite{zhou20213d} & \xmark & \xmark & 73.8 & 64.8 & 150 & 42.0 & 56.3 & 53.3 & 213 & 31.2 & 54.6 & 53.8 & 275 & 58.4 \\
 LION~\cite{vahdat2022lion} & \xmark & \xmark & 67.4 & 61.2 & 97.2 & 40.6 & 53.7 & 52.3 & 168 & 30.8 & 53.4 & 51.1 & 201 & 55.2 \\
 SPAGHETTI~\cite{hertz2022spaghetti} & \cmark & \xmark & 78.2 & 77.0 & 1530 & 529 & 72.3 & 71.0 & 581 & 284 & 70.7 & 69.0 & 5930 & 582 \\
 DiT-3D~\cite{mo2023dit} & \xmark & \xmark & 64.7 & 60.3 & 105 & 42.4 & 52.7 & \textbf{50.2} & 206 & 327 & 52.5 & 53.1 & 235 & 49.0 \\
  SALAD~\cite{koo2023salad} & \cmark & \xmark & 73.9 & 71.1 & 198 & 45.1 & 59.2 & 57.2 & 236 & 29.4 & 57.8 & 58.4 & 308 & 52.6 \\
 FrePolad~\cite{zhou2024frepolad} & \xmark & \xmark & 65.3 & 62.1 & 94.1 & 38.1 & 52.4 & 53.2 & 173 & 29.6 & 51.9 & \textbf{50.3} & 252 & 50.9 \\
\ours (ours) & \cmark & \cmark & \textbf{63.3} & \textbf{59.7} & \textbf{25.7} & \textbf{1.87} & \textbf{50.1} & 51.8 & \textbf{25.7} & \textbf{2.28} & \textbf{51.6} & 53.7 & \textbf{28.9} & \textbf{2.86} \\ 
 \bottomrule
\end{tabular}}
}
\end{table}
%%%%%%%%%%%%%%%%%%%%%%%%%%%%%%%%%%%%%%%%%%%%%%%%%%%%%%%%%%%%

\end{document}